\documentclass{article}
\usepackage{spconf,amsmath,graphicx}

\title{Writer Recognition Using Off-line Handwritten\\ Single Block Characters}
\name{Adrian Leo Hagström, Rustam Stanikzai, Josef Bigun, Fernando Alonso-Fernandez\thanks{This work has been carried out by Adrian Leo Hagström and Rustam Stanikzai in the context of their Master Thesis at Halmstad University (Engineering, Computer Science and Engineering program). Authors acknowledge the funding of the Swedish Research Council (VR), Innovation Agency (Vinnova) and Knowledge Foundation (KKS).}}
\address{School of Information Technology (ITE), Halmstad University, Sweden\\ adrhag15@student.hh.se, russta15@student.hh.se, josef.bigun@hh.se, feralo@hh.se}
\begin{document}
%
\maketitle
\begin{abstract}

Block characters are often used when filling paper forms for a variety of purposes.
We investigate if there is biometric information contained within individual digits of handwritten text. In particular, we use personal identity numbers consisting of the six digits of the date of birth, DoB.
We evaluate two recognition approaches, one based on handcrafted features that compute contour directional measurements, and another based on deep features from a ResNet50 model. 
We use 
a self-captured database of 317 individuals and 4920 written DoBs in total. 
Results show the presence of identity-related information in a piece of handwritten information as small as six digits with the DoB.
We also analyze the impact of the amount of enrolment samples, varying its number between one and ten.
Results with such small amount of data are promising. With ten enrolment samples, the Top-1 accuracy with deep features is around 94\%, and reaches nearly 100\% by Top-10. 
%
%
%
The verification accuracy is more modest, with EER$>$20\% with any given feature and enrolment set size, showing that there is still room for improvement. 
\end{abstract}
\section{Introduction}
\label{sec:intro}

Recognition of individuals using handwritten text or signature is an established topic in biometrics \cite{Hafemann17ipta_offline_signature_soa,Rehman19mta_writer_soa} because of the social and legal acceptance and the widespread use of documents and forms filled by hand.
Here, we address the challenge of automatic recognition by block characters of numbers containing the personal ID number. We use scanned images of forms, known as \textit{off-line} writer recognition, so only static information is available. This is in contraposition to \textit{on-line} recognition, where pen tablets or digitizers that capture dynamic information such as velocity, acceleration, pressure or pen inclination are used \cite{[Plamondon00]}. 
In Sweden, the personal ID number is a unique number derived from the DoB. Such unique numbers exist in nearly every country, e.g. social security number (USA), pension number (Switzerland), etc. 
Analyzing identity via handwritten material has been addressed 
using signature or free text. 
For many years, research have been devoted to finding good handcrafted features for offline writer recognition \cite{Hafemann17ipta_offline_signature_soa,Rehman19mta_writer_soa}.
The field has not remained oblivious to the deep learning explosion, with recent works employing Convolutional Networks to extract features that are used to train a traditional classifier \cite{Fiel15CAIP_WriterIDcnn,Hafemann17PR_offline_signature_verification_cnn}, or to do the classification directy end-to-end \cite{Dey17arxiv_SigNet_Siamese_offline_writer}.
%
%
However, to the best of our knowledge, using a reduced amount of individual characters or block letters has been barely addressed. 
Some related works \cite{Kutzner19electronics_writerIDtextwords,Tolosana2020tifs_BioTouchPass2_Tolosana} have a similar purpose, but using passwords written on a smartphone touch-screen, thus accounting for on-line information. 
In \cite{Kutzner19electronics_writerIDtextwords}, the authors use passwords of 8 characters. 
The best reported identification accuracy is 95.38\% with a combination of geometrical, statistical and temporal features.
%
The paper \cite{Tolosana2020tifs_BioTouchPass2_Tolosana} employs a 4-digits password. 
With a combination of Dynamic Time Warping and Recurrent Neural Networks, they achieve an Equal Error Rate of just 2.38\% with only one training sample per character.
In a similar vein to the present paper, the work \cite{[Alonso10a]} uses off-line characters segmented from written texts. However, the task is not equivalent, since it considers 62 alphanumeric classes corresponding to numbers and the entire alphabet (upper- and lowercase), and data is extracted from free texts. 
In addition, characters belonging to the same class are grouped, while the present paper keeps all digits of the ID number separated, even if they repeat.
With a database of 30 writers, the best Top-1 identification accuracy reported by \cite{[Alonso10a]} is $\sim$58\%. A bigger hit list size results in an accuracy of $\sim$87\% (Top-5) or $\sim$92\% (Top-10).

Our focus is thus to establish if a small amount of off-line handwritten evidence (just 6 digits with the DoB) contains sufficient identity information to recognize individuals. 
Since people is used to write such information frequently on paper, we hyphotesize that its writing is automatized in a way that it must be as personal as for example a signature \cite{[Fairhust97]}.
For the study, we have selected two methods, one based on contour directional features previously employed for off-line signature and writer recognition \cite{[Gilperez08],[Alonso10a]}, and another based on deep features extracted from a ResNet50 \cite{[He16]}, which is a very popular backbone for biometric recognition \cite{Chen19IS_SemisupervisedWriterID,[Cao18vggface2]}. 
Our results show a clear indication of recoverable identity related information within the DoB digits handwritten in boxes.

\begin{figure}[t]
\centering
\includegraphics[width=0.45\textwidth]{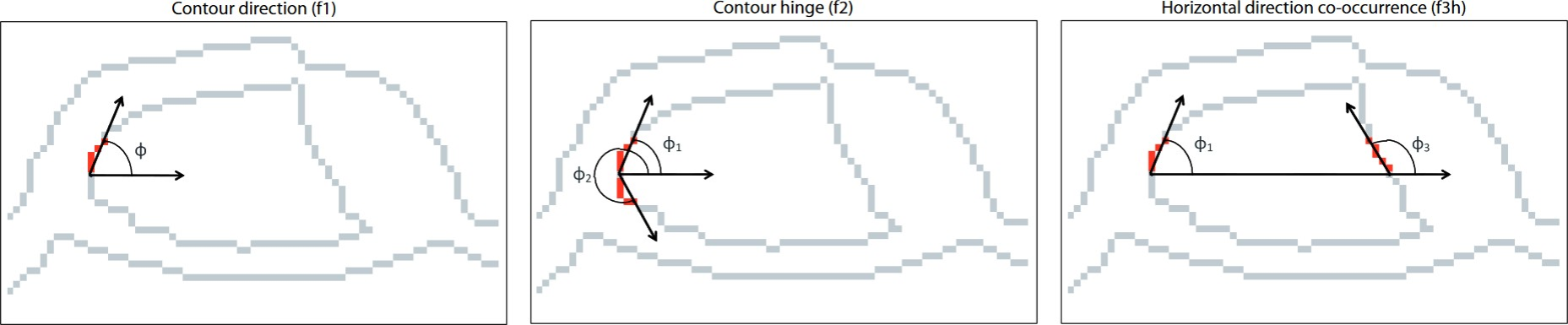}
\caption{Graphical example of handcrafted features.}
\label{fig:fig1-pdfs}
\end{figure}

\section{Recognition Systems}
\label{sec:system}

We employ the handcrafted contour features 
used for signature and handwritting recognition in \cite{[Gilperez08],[Alonso10a]}.
The features were originally presented in \cite{[Bulacu07]}.
They are: direction (f1), hinge (f2) and directional co-ocurrence (f3h, f3v). 
A graphical description of its extraction is shown in Figure~\ref{fig:fig1-pdfs}.
These features start from the internal and external contours of connected components. 
The direction (f1) encodes the orientation $\phi$ of local contour fragments. The hinge (f2) jointly captures the orientations $\phi_1$ and $\phi_2$ of two contour fragments attached to a common end pixel, thus encoding the local curvature.
%
The directional co-ocurrence (f3h, f3v) encodes two joint orientations, but at the ends of run-lengths on the background, thus capturing the overall shape of the character (e.g. an elongated vs. round character).
The algorithm runs over the contour to build the histograms of the different features. When finished, they are converted to probability distribution functions (PDFs).
To compute the similarity between two different PDFs, the $\chi ^2$ distance is used \cite{[Bulacu07]}.

For deep learning experiments, we choose a ResNet50 \cite{[He16]} due to its high performance and popularity in a wide variety of biometric applications \cite{Chen19IS_SemisupervisedWriterID,[Cao18vggface2]}.
ResNet introduced the concept of residual connections to ease the training of CNNs, allowing much deeper networks. The key idea is to make available the input of a lower layer to a higher layer, bypassing intermediate ones. 
This way, gradients flow through the network directly, without passing through non-linear activation functions, mitigating the problem of vanishing or exploding gradients.
The ResNet50 model is modified, so that after the Global Average Pooling, we add a fully-connected layer of 512 elements, followed by a softmax output layer with 317 classes. 
The input layer is also modified to accept images of size 35$\times$138.
The input image size and number of classes is defined by the database and experimental protocol employed, as described in the next section. 

\begin{figure}[t]
\centering
\includegraphics[width=0.45\textwidth]{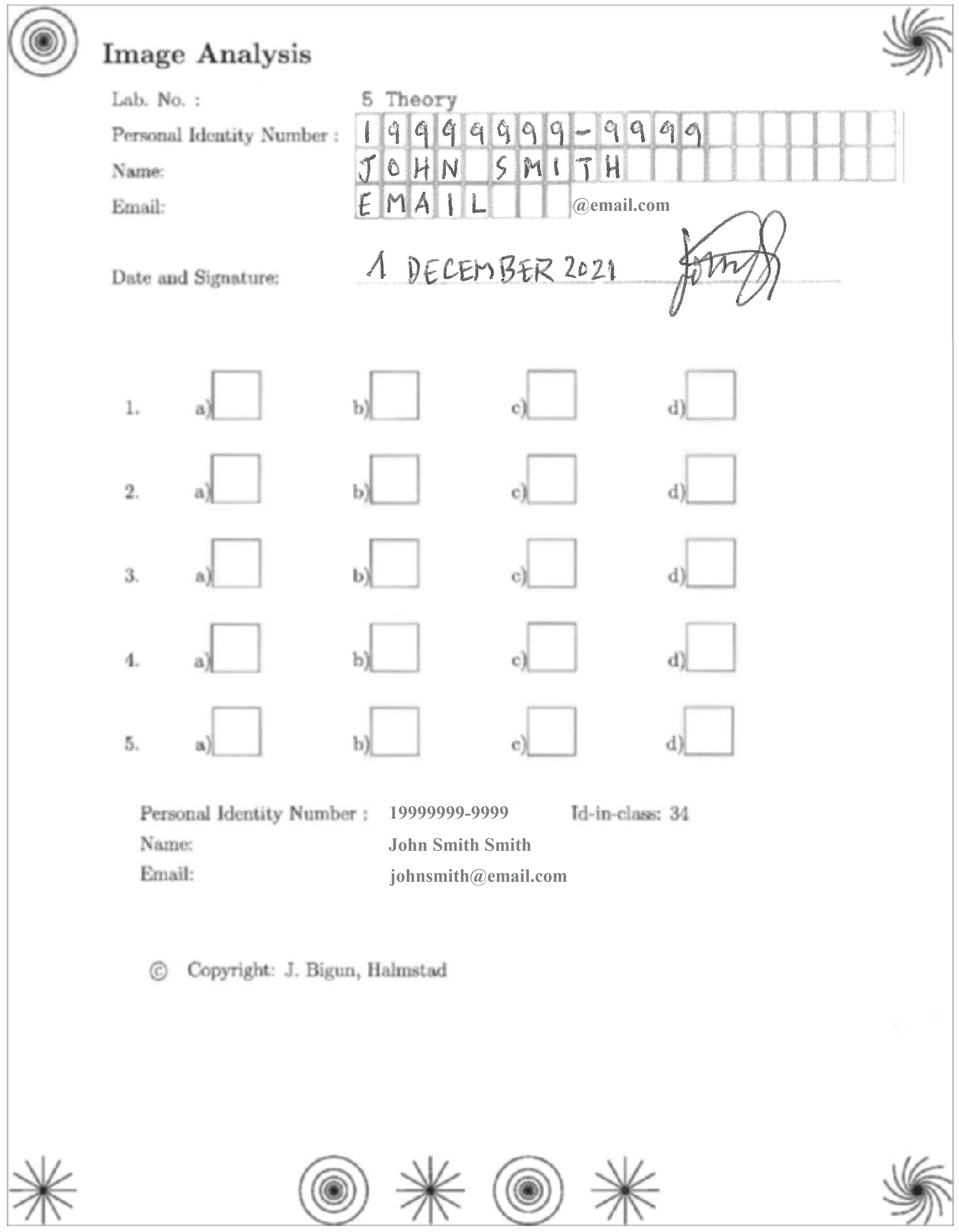}
\caption{Form where digits and other characters are filled.}
\label{fig:fig4-blocksexample}
\end{figure}

\section{Database and Protocol}
\label{sec:dbprotocol}

Data is obtained from answer sheets of student quizzes carried out at Halmstad University. 
The sheets have boxes at the top where students fill the ID number, name, email, etc. (Figure~\ref{fig:fig4-blocksexample}).
After scanning, the spiral markers shown in the corners allow to align the page (compensating rotation, scale, etc.) and to find the boxes that contain the student information and answers, since the boxes are at predefined positions \cite{Alonso22educon_spirals}. 
For this work, only the ID number is used. In Sweden, it has the form `YYYYmmdd-aaaa', where the first 8 characters are the date of birth (DoB), and the last 4 characters are specific of each individual. 
For privacy assurance during data handling in this work, we only use the DoB, since the last 4 digits would allow identification of the individual by just reading the images. Likewise, name and email data are also concealed.
In addition, some people abbreviate the DoB to six digits 
(e.g. 99mmdd instead of 1999mmdd), so we only employ the last 6 digits of the DoB.
If the feasibility of the method is demonstrated with just six digits, then it can be easily extended to consider the full ten/twelve digits of the ID number, as well as the name and other form details. 

\begin{figure}[t]
\centering
\includegraphics[width=0.45\textwidth]{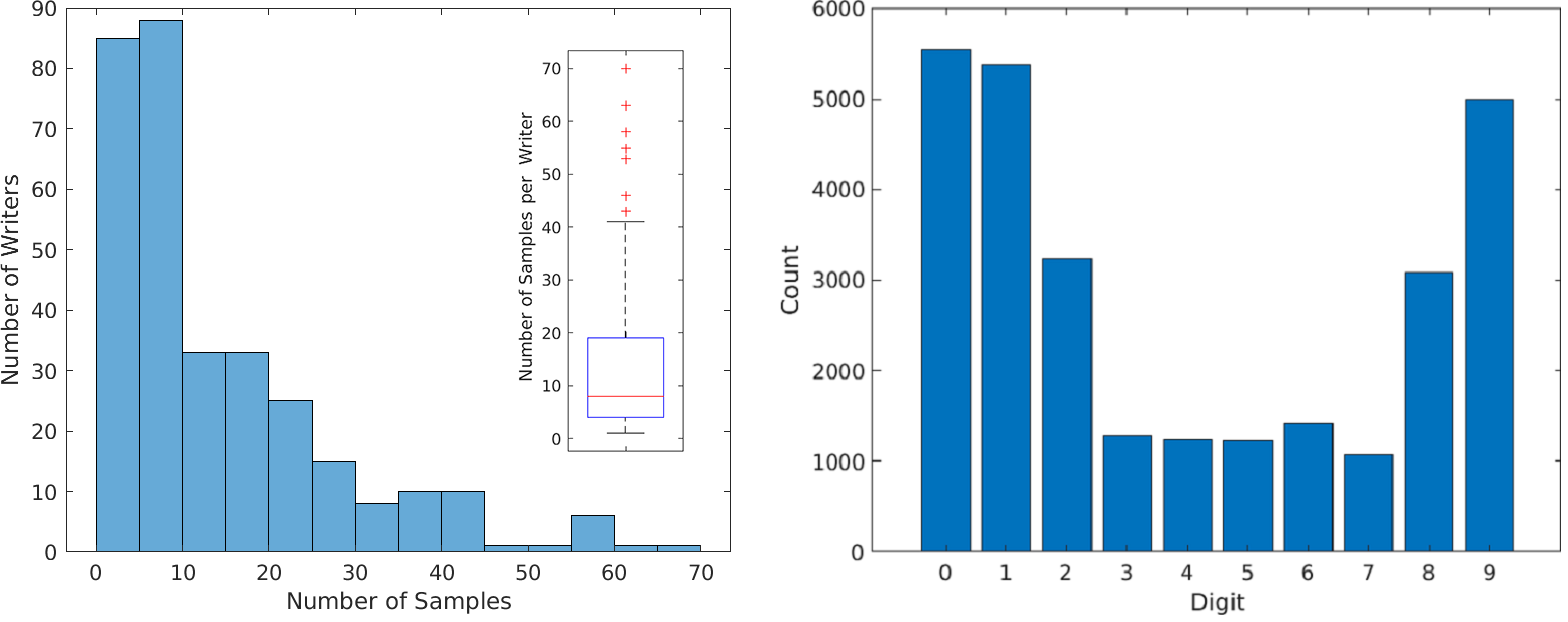}
\caption{Statistics of the database. Left: histogram of samples (DoBs) per writer. Right: count of images per digit.}
\label{fig:fig5-6-dbstats}
\end{figure}

The gathered database has 317 people with 4920 written DoBs in total. 
The amount of samples per person varies, see Figure~\ref{fig:fig5-6-dbstats}, left (1 sample = 1 set of 6 digits). Some people have over 80 samples, but the majority have 5-20 samples, with a median of 8 and an average of 13.6 samples per writer.
Data has been obtained across the years via regular weekly quiz examinations in two master courses at the mentioned university (something that can be deducted by examining Figure~\ref{fig:fig4-blocksexample}). Some students take both courses, while other students take only one, and the weekly quizzes are voluntary. This explains the unbalance of samples per person.
Figure~\ref{fig:fig5-6-dbstats}, right, shows the count of images per digit in the database, with a preference for digits like 0, 1, 8, 9. This is due to prevalence of 8x, 9x, 0x in the year digits, or 0x, 1x in the month digits. 
The digits of each sample are then extracted to form six 35$\times$27 pixel sub-images, and manually labelled to 0-9 as well. 
The box containing each digit is cut so no boundary remains in any side of the image.
To remove scan noise, the Otsu method is then applied to binarize the images. %
%

We carry out both identification and verification experiments.
The impact of the amount $N$ of samples for enrolment is also studied ($N$=1 to 10).
Identification 
is done by taking $N$ samples randomly from each person for enrolment, and using the remaining samples for testing. 
%
Given a test sample, it is compared against every enrolment sample, producing a sorted list of identities from the best to the worst match. Results are reported in the form of Top-$M$ or Rank-$M$ measure, which gives the probability that the correct identity is within the first $M$ matches.
Regarding verification experiments, genuine matches are done by taking the first $N$ samples from each person for enrolment, and using the remaining samples for testing. If the writer has more than 2$\times N$ samples, then the next $N$ samples not yet used are taken for enrolment, and the rest for testing. This is repeated for writers with more than 3$\times N$ samples, etc. until all samples are exhausted. People with less than $N$ samples are not used since they do not have enough data for client tests. 
Impostor matches are done by creating a `pseudo-forgery' for every validation sample by using random digit-entries from different writers to build the same personal number as the current one being tested.

\begin{figure}[t]
\centering
\includegraphics[width=0.4\textwidth]{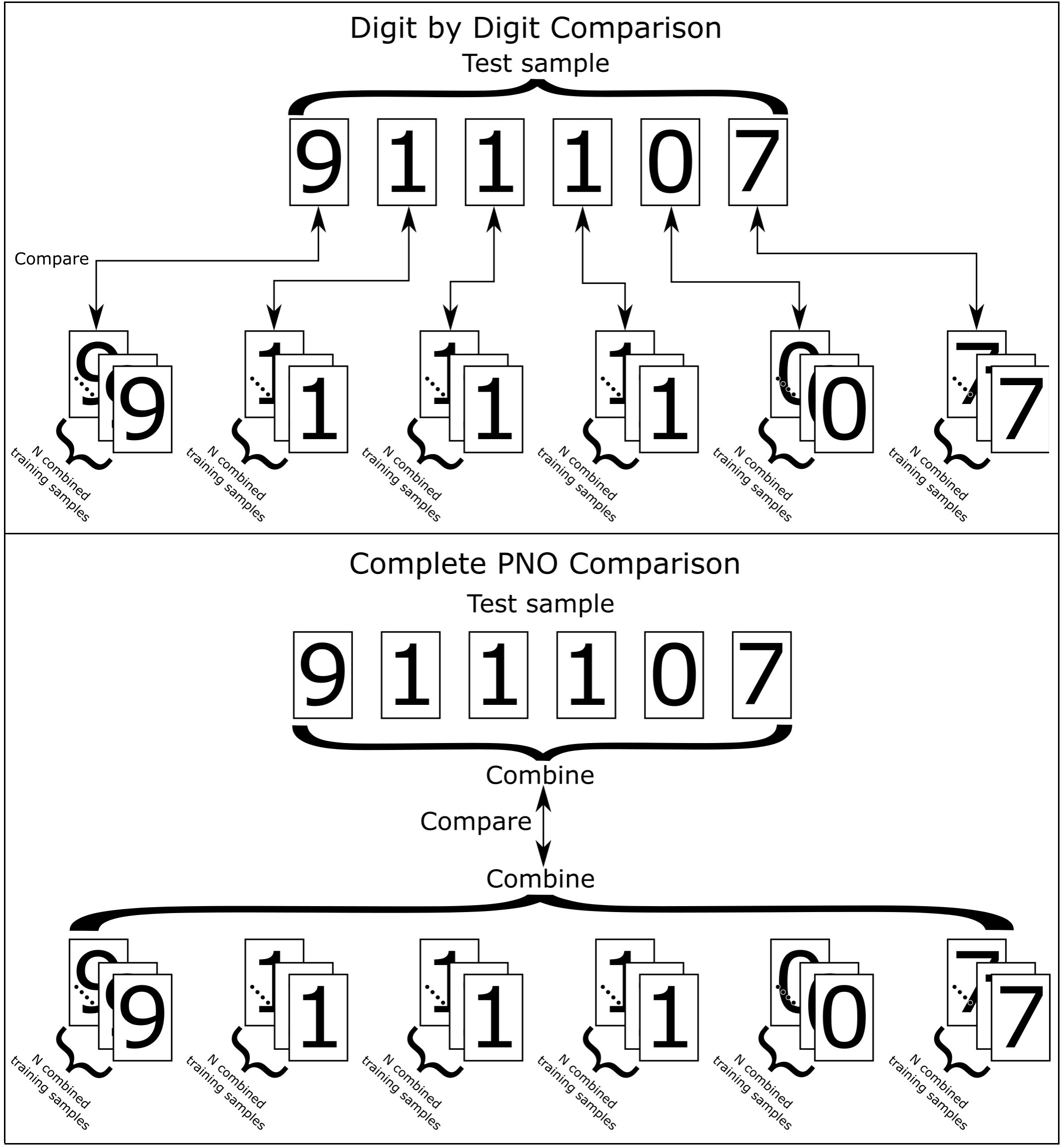}
\caption{Comparison options using handcrafted features.}
\label{fig:fig2-samples-comparison}
\end{figure}

When comparing samples using handcrafted features, we do it in two ways (Figure~\ref{fig:fig2-samples-comparison}). The $N$ feature vectors of the enrolment samples are first combined digit-wise by averaging. 
Then, we either: $i$) compute digit by digit the distance between enrolment and test samples, resulting in six distance values, which are then averaged (top part of the figure); or $ii$) average the six feature vectors of all digits, and then obtain a single distance value between the enrolment and test samples (bottom part). 
%
%
%
With regards to the CNN, it is first trained for biometric identification to classify which of the 317 writers wrote an input sample. 
The six 35$\times$27 sub-images (individual digits) of a data sample are concatenated into a single 35$\times$138 image, which is used as input to the CNN. 
The trained CNN models are then used for feature extraction, using the output of the next to last layer of the network (of size 512) as feature vector of a given sample.
The metric to compare two CNN feature vectors is the Euclidean distance.
%
%
Given the limited amount of data, the network is initialized using ImageNet weights. When training the CNN, we either freeze the network (except the fully-connected layers) or allow that the entire network is updated.
The frozen model is trained for 10 epochs, and the best epoch w.r.t. validation loss is saved.
To combat over-fitting, the saved frozen model is used as initialization of the unfrozen model, which is trained for 20 epochs, again saving the model of the best epoch. 
Because the optimal learning rate for each model may vary, the hyper-parameter optimization framework Optuna is used \cite{Akiba19SIGKDD_OptunaFramework}. 
Data augmentation could also be beneficial, so to improve generalization ability and prevent that the model just learn to read the order of the handwritten digits, we shuffle the order of the digits when composing the 35$\times$138 image. This is done for all the training and validation samples at the beginning of each epoch.
%
%
%

\begin{figure*}[htb]
\centering
\includegraphics[width=0.95\textwidth]{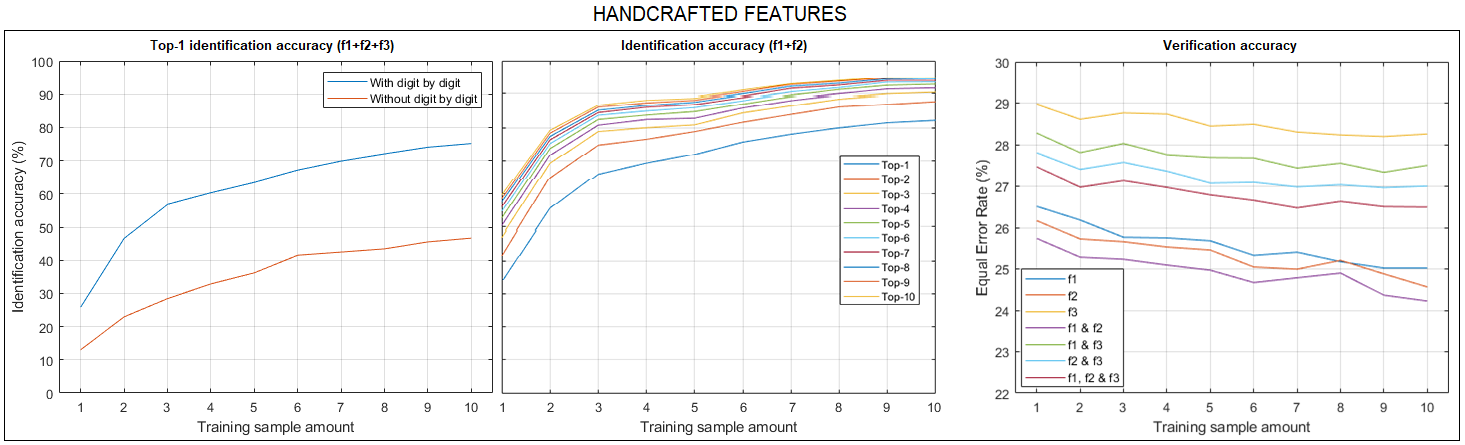}
\caption{Handcrafted features: identification and verification accuracy using a different amount of training samples.}
\label{fig:fig12-14-15-handcrafted}
\end{figure*}

\begin{figure*}[htb]
\centering
\includegraphics[width=0.95\textwidth]{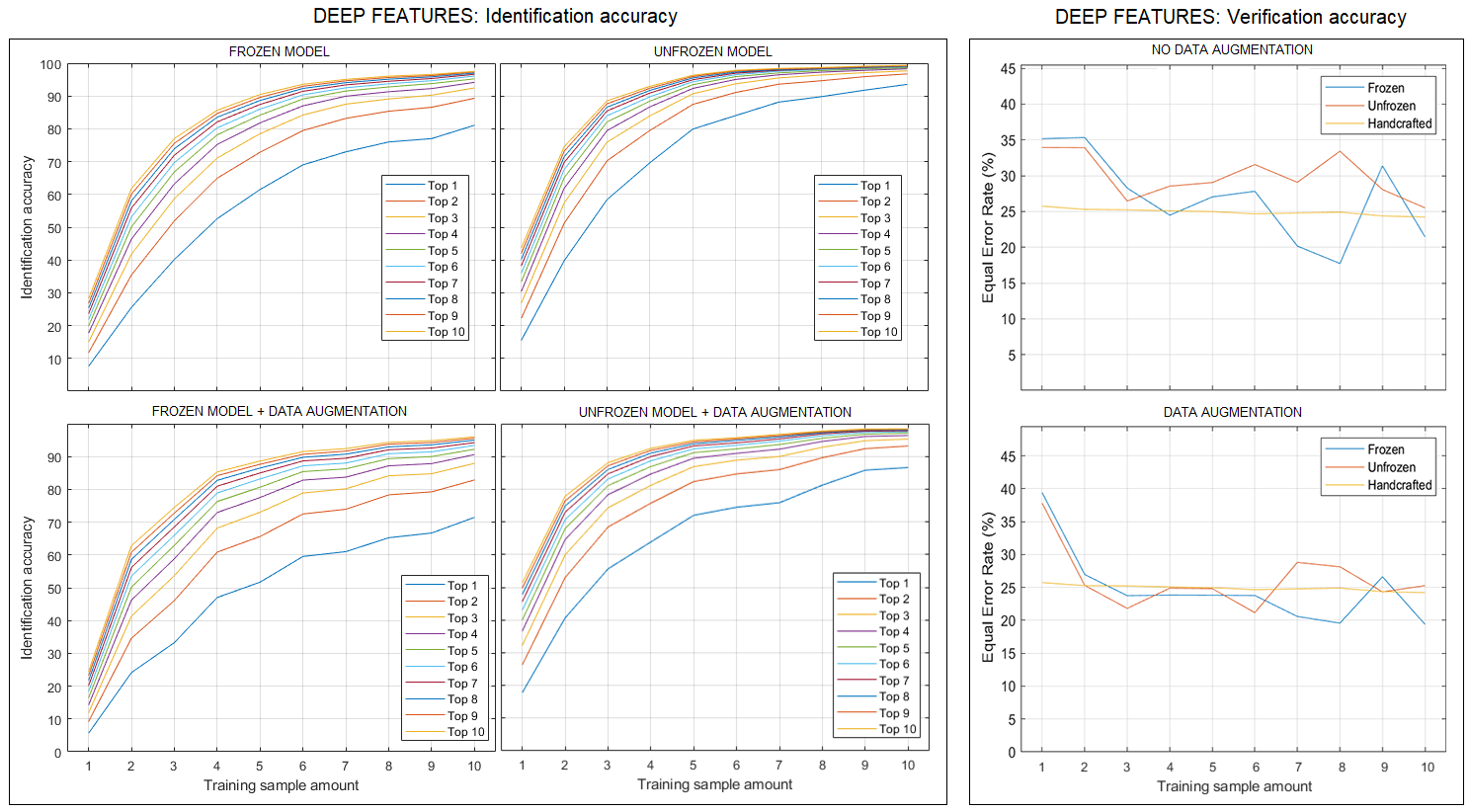}
\caption{Deep features: identification and verification accuracy using a different amount of training samples.}
\label{fig:fig19-20-22-deep}
\end{figure*}

\begin{figure}[htb]
\centering
\includegraphics[width=0.4\textwidth]{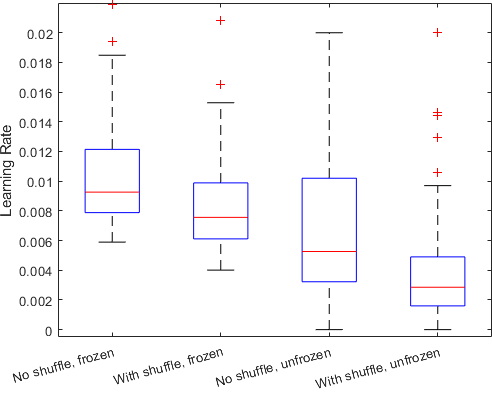}
\caption{Final learning rates of the deep models from Optuna. Each box-plot represents 100 learning rates.}
\label{fig:fig32a-learningrates}
\end{figure}

\section{Results}
\label{sec:results}

Figure~\ref{fig:fig12-14-15-handcrafted} reports the results with handcrafted features using different amounts of enrolment (training) samples. 
The left plot shows the Top-1 identification accuracy when using digit by digit distance (Figure~\ref{fig:fig2-samples-comparison}, top) or not (bottom).
%
The plot refers to the combination of f1, f2 and f3 features (averaging the distance between two samples given by each feature).
The digit by digit option is shown to be much better, so it will be used for the rest of the paper.
This is very likely due to being able to compare separately how each digit is written, since different digits have different contours. 
In addition, writers may have different ID numbers, so allowing comparison of separate digits will likely produce higher distances in digits that are not equal in both writers. 
Even if two writers would have the same DoB, we hypothesize that keeping digits separated allows to capture differences in the process of writing each digit one after the other, which can be expected to be different for each person. 
If feature vectors of all digits are averaged, such benefits are diluted. 
This result is also observed in \cite{[Alonso10a]}, where keeping classes separated is shown to be better than averaging the feature vectors of all classes first, and then computing a single distance measure.

We have next evaluated the identification accuracy of the handcrafted features, including their combinations. The best results (up to Top-10) are given by combination of f1 and f2, whose plot is given in the center of Figure~\ref{fig:fig2-samples-comparison} (all other options not shown due to space). 
Apart from the obvious increase in performance with the addition of enrolment samples, the sharpest increase is shown with the use of two or three. Using four or more samples does increase as well, but to a lesser extent. 
The same can be said about the Top-$M$ performance, which increases sharply up to $M$=3 and then, the increase is more modest.
In accuracy terms, it is relevant for example that with eight or more samples, the Top-1 accuracy is higher than 80\%, or that the Top-10 accuracy reaches 90\% with just five-six samples.  
If such amount of enrolment samples would be available, it is striking the good identification accuracy numbers observed with just the six digits of the DoB.
On the other hand, verification results (Figure~\ref{fig:fig12-14-15-handcrafted}, right) does not show so bright results. 
We give results of different features (including combinations), showing here as well that the best result is given by f1+f2. Feature f3 (combination of f3v+f3h) shows the worst accuracy, and it does not seem to contribute to fusion, since it actually worsens the performance of other features when combined. 
On the other hand, features f1 and f2 are shown to be powerful, and even better when combined. 
The absolute values of the EER for f1+f2 ranges from $\sim$25.7\% (one enrolment sample) to $\sim$24.2\% (ten enrolment samples). 
Even if such results are modest, we do not have a baseline to compare with, so we see them as a starting point for the use of the handwritten DoB for recognition.

Experiments with the deep learning model are then shown in Figure~\ref{fig:fig19-20-22-deep}. 
The left and center columns show the identification results with the frozen and unfrozen CNN, respectively. 
Also, the first row is without data augmentation, whereas the second row is with data augmentation by shuffling the digit order.
A significant improvement is observed with the unfrozen model (center vs. left column), but the same cannot be said about the data augmentation method (second vs. first row).
With the best model (unfrozen, no data augmentation), results are observed to be worse than the handcrafted features if few enrolment samples are available. For example, with one enrolment sample, Top-10 with handcrafted features reaches 60\%, whereas it is of just 44\% with CNN features. With two enrolment samples, Top-10 is 80\% (handcrafted) vs. 75\% (CNN).
On the other hand, if more enrolment samples are available, the CNN features surpasses the handcrafted ones. With 5 enrolment samples, the CNN already reaches Top-10=95\%, and 98\% with seven samples. 
Also, with eight samples, the Top-1 is 90\%. 

On the right column of Figure~\ref{fig:fig19-20-22-deep}, we also provide verification results of the CNN models. The yellow curve shows the results with handcrafted features (f1+f2) for comparison.
Here, data augmentation does actually produce a performance improvement. On the other hand, unfreezing the model does not given an improvement (and it even worsens in some cases). 
It is also noticeable a massive drop in EER after combining just two samples for enrolment, although the inclusion of more samples seems to stop the improvement tendency. 
Such lack of improvement with more data and the the somewhat oscillating results of the CNN models might be attributed to over-fitting due to few data available, but this would need to be confirmed with extra data and experiments.
It can be also seen that, as in the identification experiments, if there is few enrolment data (just one sample here), handcrafted features give much better performance. 
However, the use of just two or three samples with the CNN already matches or beats the handcrafted features, something not seen in the identification experiments. 
All in all, the CNN appear to do a little bit better compared with the handcrafted features (EER of 20-24\% vs. 24\%). Still, we consider these results as a starting point, with plenty of room for improvement.

The learning rates of the CNN models from Optuna are shown in Figure~\ref{fig:fig32a-learningrates}.
They have a pattern where the unfrozen models tend to have higher variance, whereas the shuffle data augmentation produces smaller learning rates. 
This is likely due to a regularization effect caused by the data augmentation, and the unfrozen model probably leading to over-fitting.

\section{Conclusions}
\label{sec:conclusions}


Our goal is to find whether there is any identity related information in the images of handwritten digits of personal ID numbers.
Many documents containing personal information such as name, date of birth (DoB), ID number, signature, etc. are filled by hand and accepted as a legal proof in a variety of environments.
In particular, we use 
the six digits of the DoB, which is used to build the personal ID number in the country where we have carried out this study.
Since data consists of images, the challenge is categorized as 
%
%
off-line recognition, which is more complex compared to on-line operation, since there is not dynamic information available. 
On the other hand, infrastructure for on-line acquisition is not always available, so the development of off-line methods is highly relevant. 

In this work, we employ handcrafted 
features previously used for signature and writer recognition which encode contour direction, curvature, or character shape. We also extract character features using a ResNet50 backbone. 
Experimental results are given using a self-acquired database of 317 writers, with 4920 written DoBs. 
Accuracy of the features is assessed both in biometric identification and verification mode. We also evaluate the impact of the amount of samples (images with the DoB) used for enrolment, varying its amount between one and ten.
In general, the CNN features surpass the handcrafted features, unless the amount of enrolment samples is very low (one or two). 
With only one enrolment sample, the Top-10 identification accuracy with handcrafted features is 60\%, and increases to 80\% with two samples (44\% and 75\% respectively with the CNN). 
On the other hand, if we allow five enrolment samples, the CNN reaches Top-1=80\%, and 90\% with eight samples, while handcrafted features remain at 72\% and 80\% respectively. 
Adding more samples or considering a bigger hit list size allows to obtain even better results (e.g. Top-10=98\% with seven samples).
Regarding verification results, the numbers are more modest, with EER$>$20\% 
for any amount of enrolment samples and feature type.
Also, the handcrafted features are more consistent to changes in the amount of samples, and more resilient to the availability of few data. Its performance with one sample (EER$\sim$25.8\%) improves steadily as we add more samples, reaching $\sim$24.2\% with ten samples.
On the other hand, the CNN features start at a very high value (EER$>$34\% with one sample), and show a significant EER drop as we use two or three samples, but the inclusion of more data does not imply a better EER. The EER curve shows a somehow oscillating behaviour as well. 

As a way to improve performance, and solve the oscillating behaviour of the CNN, we are looking ways to augment the available data. 
One option is semi-supervised data augmentation using segmented characters from other databases \cite{[Hull94]}, as proposed for example in \cite{Chen19IS_SemisupervisedWriterID}. 
Generative approaches to create synthetic image data, e.g. based on Generative Adversarial Networks \cite{[Yapici21paa_OfflineSignatureAugmentation]} or dynamics of on-line data \cite{Galbally2015_PR_OnOfflineSignSynthesis}, will be also investigated. 
Another avenue would be to train the network on individual digits instead of using an image with the whole ID number. This would allow to compare features digit by digit, an approach that has shown to be better with the handcrafted features.
It could also be that the chosen CNN is too complex for the task or the amount of available data.
Other CNN backbones with less learnable parameters such as SqueezeNet \cite{[Iandola16SqueezeNet]} or MobileNet \cite{[Sandler18mobilenetv2]} will thus be looked at.
Given the different behaviour of the features depending on the amount of enrolment data, specially in verification, fusion of handcrafted and deep features is another path to exploit their complementarity and improve performance as well \cite{[Hernandez18]}.
%


\bibliographystyle{IEEEbib}

\small

\begin{thebibliography}{10}

\bibitem{Hafemann17ipta_offline_signature_soa}
L.~G. Hafemann et al.,
\newblock ``Offline handwritten signature verification — literature review,''
\newblock in {\em Proc. IPTA}, 2017.

\bibitem{Rehman19mta_writer_soa}
A.~Rehman et al.,
\newblock ``Writer identification using machine learning approaches: A
  comprehensive review,''
\newblock {\em Multim. Tools Appl.}, v. 78, 2019.

\bibitem{[Plamondon00]}
R.~Plamondon, S.N. Srihari,
\newblock ``On-line and off-line handwriting recogn: A comprehensive
  survey,''
\newblock {\em IEEE TPAMI}, 22(1), 2000.

\bibitem{Fiel15CAIP_WriterIDcnn}
S.~Fiel, R.~Sablatnig,
\newblock ``Writer identification and retrieval using a cnn,''
\newblock {\em Proc. CAIP}, LNCS-9256, 2015

\bibitem{Hafemann17PR_offline_signature_verification_cnn}
L.~G. Hafemann et al.,
\newblock ``Learning features for offline handwritten signature verif. using deep cnns,''
\newblock {\em Patt. Recogn.}, 2017.

\bibitem{Dey17arxiv_SigNet_Siamese_offline_writer}
S.~Dey et al.,
\newblock ``Signet: Convolutional siamese network for writer independent
  offline signature verif.,''
\newblock {\em CoRR}, vol. abs/1707.02131, 2017.

\bibitem{Kutzner19electronics_writerIDtextwords}
T.~Kutzner et al.,
\newblock ``Writer identification using handwritten cursive texts and single
  character words,''
\newblock {\em Electronics}, 8(4), 2019.

\bibitem{Tolosana2020tifs_BioTouchPass2_Tolosana}
R.~Tolosana et al.,
\newblock ``Biotouchpass2: Touchscreen password biometrics using time-aligned
  recurrent neural networks,''
\newblock {\em IEEE TIFS}, vol.
  5, 2020.

\bibitem{[Alonso10a]}
F.~{Alonso-Fernandez} et al.,
\newblock ``Exploiting character class information in forensic writer
  identification,''
\newblock {\em Proc IWCF}, LNCS-6540, 2010.

\bibitem{[Fairhust97]}
{M. C.} Fairhurst,
\newblock ``Signature verification revisited: promoting practical exploitation
  of biometric technology,''
\newblock {\em IET Electronics and Communication Engineering Journal}, vol. 9, 1997.

\bibitem{[Gilperez08]}
A.~Gilperez, F.~{Alonso-Fernandez} et al.,
\newblock ``Off-line signature verification using contour features,''
\newblock {\em Proc ICFHR},
  2008.

\bibitem{[He16]}
K.~He et al.,
\newblock ``Deep residual learning for image recognition,''
\newblock in {\em Proc CVPR}, June 2016, pp. 770--778.

\bibitem{Chen19IS_SemisupervisedWriterID}
S.~Chen et al.,
\newblock ``Semi-supervised feature learning for improving writer
  identification,''
\newblock {\em Information Sciences}, vol. 482, 2019.

\bibitem{[Cao18vggface2]}
Q.~{Cao} et al.,
\newblock ``Vggface2: A dataset for recognising faces across pose and age,''
\newblock {\em Proc. Intl Conf Automatic Face \&
  Gesture Recogn. FG}, 2018.

\bibitem{[Bulacu07]}
M.~Bulacu, L.~Schomaker,
\newblock ``Text-independent writer identification \& verif. using
  textural \& allographic features,''
\newblock {\em IEEE TPAMI}, vol. 29, 2007.

\bibitem{Alonso22educon_spirals}
F.~{Alonso-Fernandez}, J.~Bigun,
\newblock ``Continuous examination by automatic quiz assessment using spiral
  codes and image processing,''
\newblock {\em Proc EDUCON (accepted)}, 2022.

\bibitem{Akiba19SIGKDD_OptunaFramework}
T.~Akiba et al.,
\newblock ``Optuna: A next-generation hyperparameter optimization framework,''
\newblock in {\em Proc SIGKDD}, 2019.

\bibitem{[Hull94]}
{J.J.} Hull,
\newblock ``A database for handwritten text recognition research,''
\newblock {\em IEEE TPAMI}, vol.
  16, no. 5, pp. 550--554, May 1994.

\bibitem{[Yapici21paa_OfflineSignatureAugmentation]}
M.M. Yapıcı et al.,
\newblock ``Deep learning-based data augmentation method and signature
  verification for offline handwritten signature,''
\newblock {\em Pattern Analysis and Applications}, vol. 24, 2021.

\bibitem{Galbally2015_PR_OnOfflineSignSynthesis}
J.~Galbally et al.,
\newblock ``On-line signature recognition through the combination of real
  dynamic data and synthetically generated static data,''
\newblock {\em Pattern Recognition}, vol. 48, 2015.

\bibitem{[Iandola16SqueezeNet]}
F.~N. Iandola et al.,
\newblock ``Squeezenet: Alexnet-level accuracy w. 50x fewer parameters and
  {\textless}1mb model size,''
\newblock {\em CoRR}, vol. abs/1602.07360, 2016.

\bibitem{[Sandler18mobilenetv2]}
M.~{Sandler} et al.,
\newblock ``Mobilenetv2: Inverted residuals and linear bottlenecks,''
\newblock in {\em Proc CVPR}, 2018, pp. 4510--4520.

\bibitem{[Hernandez18]}
K.~Hernandez-Diaz, F.~Alonso-Fernandez, J.~Bigun,
\newblock ``Periocular recognition using {CNN} features off-the-shelf,''
\newblock in {\em Proc BIOSIG},
  Sep. 2018.

\end{thebibliography}

\end{document}